\documentclass[runningheads,a4paper]{llncs}

\usepackage{amssymb}
\usepackage{graphicx}

\usepackage{url}
\urldef{\mailsa}\path|{alfred.hofmann, ursula.barth, ingrid.haas, frank.holzwarth,|
\urldef{\mailsb}\path|anna.kramer, leonie.kunz, christine.reiss, nicole.sator,|
\urldef{\mailsc}\path|erika.siebert-cole, peter.strasser, lncs}@springer.com|    

\usepackage[utf8]{inputenc} 
\usepackage[T1]{fontenc}    
\usepackage{hyperref}       
\usepackage{url}            
\usepackage{booktabs}       
\usepackage{amsfonts}       
\usepackage{nicefrac}       
\usepackage{microtype}      

\usepackage{amsmath,amssymb} 
\usepackage{color}
\usepackage{amsmath}
\usepackage{amssymb}
\usepackage{algorithm}
\usepackage{algorithmicx}
\usepackage{algpseudocode}
\usepackage{algpascal}

\usepackage{mathrsfs}
\usepackage{soul}
\usepackage{multirow}
\usepackage{comment}
\usepackage{xfrac}
\usepackage{rotating}
\usepackage{hhline}
\usepackage[table]{xcolor}
\usepackage{mathtools}
\usepackage{wrapfig}

\usepackage{subcaption}

\usepackage{pgfplots}
\usepackage{pgfplotstable}
\usepackage{filecontents,pgfplots}
\usepackage{tikz}
\usetikzlibrary{arrows,calc,shapes,snakes,positioning,matrix,arrows,decorations.pathmorphing,decorations.text}
\usepackage{sci}

\newcommand\blfootnote[1]
{%
\begingroup
\vspace{-5pt}
\renewcommand\thefootnote{}
\footnotetext{#1}%
\endgroup
}

\begin{document}
\mainmatter  

\title{Confounder-Aware Visualization of ConvNets}

\author{Qingyu Zhao$^{*,}$\inst{1}, Ehsan Adeli$^{*,}$\inst{1}, Adolf Pfefferbaum \inst{1,2}, \\ Edith V. Sullivan\inst{1}, \mbox{Kilian M. Pohl\inst{1,2}}}
\institute{School of Medicine, Stanford University, Stanford, USA \and Center of Health Sciences, SRI International, Menlo Park, USA}

\maketitle

\begin{abstract}
With recent advances in deep learning, neuroimaging studies increasingly rely on convolutional networks (ConvNets) to predict diagnosis based on MR images. To gain a better understanding of how a disease impacts the brain, the studies visualize the salience maps of the ConvNet highlighting voxels within the brain majorly contributing to the prediction. However, these salience maps are generally confounded, i.e., some salient regions are more predictive of confounding variables (such as age) than the diagnosis. To avoid such misinterpretation, we propose in this paper an approach that aims to visualize confounder-free saliency maps that only highlight voxels predictive of the diagnosis. The approach incorporates univariate statistical tests to identify confounding effects within the intermediate features learned by ConvNet. The influence from the subset of confounded features is then removed by a novel partial back-propagation procedure. We use this two-step approach to visualize confounder-free saliency maps extracted from synthetic and two real datasets. These experiments reveal the potential of our visualization in producing unbiased model-interpretation. 
\end{abstract}


\section{Introduction}
\blfootnote{* Equal contribution by Dr. Zhao and Dr. Adeli}\blfootnote{Source code: github.com/QingyuZhao/Confounder-Aware-CNN-Visualization.git}The development of deep-learning technologies in medicine is advancing rapidly \cite{topol2019}. Leveraging labeled big data and enhanced computational power, deep convolutional neural networks have been applied in many neuroscience studies to accurately classify patients with brain diseases from normal controls based on their MR images  \cite{topol2019,Esmaeilzadeh2018}. State-of-the-art saliency visualization techniques are used to interpret the trained model and to visualize specific brain regions that significantly contribute to the classification \cite{Esmaeilzadeh2018}. The resulting saliency map therefore provides fine-grained insights into how the disease may impact the human brain.  

Despite the promises of deep learning, there are formidable obstacles and pitfalls \cite{topol2019,he2019}. One of the most critical challenges is the algorithmic bias introduced by the model towards confounding factors in the study \cite{pourhoseingholi2012control}. A confounding factor (or confounder) correlates with both the dependent variable (group label) and independent variable (MR image) causing spurious association. For instance, if the age distribution of the disease group is different from that of the normal controls, age might become a potential confounder because one cannot differentiate whether the trained model characterizes neurodegeneration caused by the disease or by normal aging.

Since the end-to-end training scheme disfavors any additional intervention, controlling for confounding effects in deep learning is inherently difficult. This often leads to misinterpretation of the trained model during visualization: while some salient regions correspond to true impact of the disease, others are potentially linked to the confounders. In this paper, we present an approach that identifies confounding effects within a trained ConvNet and removes them to produce confounder-free visualization of the model. The central idea is first to detect confounding effects in each intermediate feature via univariate statistical testing. Then, the influence of confounded features is removed from the saliency map by a novel ``partial back-propagation'' operation, which can be intuitively explained by a chain-rule derivation on voxelwise saliency scores. This operation is efficiently implemented with a model refactorization trick. We apply our visualization procedure to interpret ConvNet classifiers trained on a synthetic dataset with known confounding effects and on two real datasets, i.e., MRIs of 345 adults for analyzing Human Immunodeficiency Virus (HIV) effects on the brain and MRIs of 674 adolescents for analyzing sexual dimporphsim. In all three experiments, our visualization shows the potential in producing unbiased saliency maps compared to traditional visualization techniques.

\section{Confounder-Aware Saliency Visualization}
We base our approach on the saliency visualization proposed in \cite{Simonyan2013DeepIC}. Given an MR image $\mathcal{I}$ and a trained ConvNet model, saliency visualization produces a voxel-wise saliency map specific to $\mathcal{I}$ indicating important regions that strongly impact the classification decision. Without loss of generality, we assume a ConvNet model is trained for a binary classification task (pipeline generalizable to multi-group classification and regression), where the prediction output is a continuous score $s \in [0,1]$. Then, the saliency value at voxel $v$ is computed as the partial derivative $|\partial s / \partial \mathcal{I}_v|$. Intuitively, it quantifies how the prediction changes with respect to a small change in the intensity value at voxel $v$. Computationally, this quantity can be computed efficiently using back-propagation. 

As discussed, when the ConvNet is confounded, some salient regions may actually contribute to the prediction of confounding variables rather than the group label. To address this issue, we propose a two-step approach to remove confounding effects from the saliency map enabling an unbiased interpretation of a trained ConvNet. To do this, we assume that a typical ConvNet architecture is composed of an encoder and a predictor. The encoder contains convolutional layers (including their variants and related operations such as pooling, batch normalization and ReLU) that extract a fixed-length feature vector $\boldsymbol{f}_i \in \mathbb{R}^M=[f_i^1,...,f_i^M]$ from the $i^{th}$ training image. The predictor, usually a fully connected network, takes the $M$ features as input and produces a prediction score $s_i$ for image $i$. To disentangle confounding effects from the saliency map, we propose in Section 2.1 to first test each of the $M$ features separately for confounding effects using a general linear model (GLM). Next, the influence from the subset of features with significant confounding effects can be removed from the saliency map by performing a novel partial back-propagation procedure based on an intuitive chain-rule derivation (Section 2.2).

\begin{figure}[!t]
\centering
\includegraphics[width=0.95\linewidth]{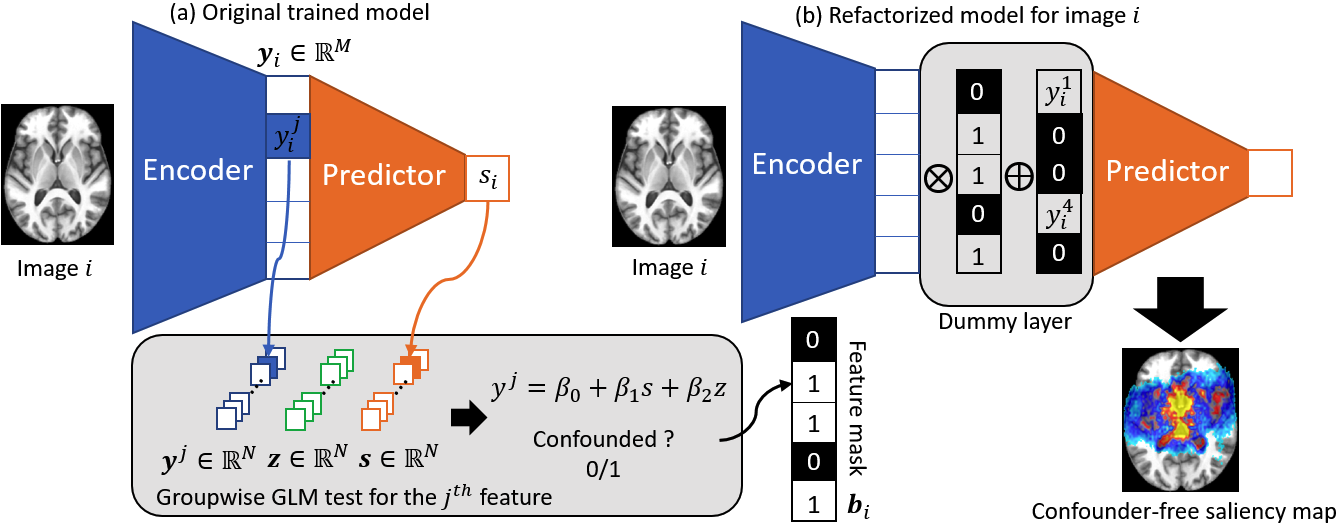}
\caption{Our confounder-aware visualization is composed two steps: (a) A GLM test is performed on each individual feature collected over all training images to detect confounding effects. (b) For each image, the model is refactorized to fix the value of confounded features, thereby enabling a partial back-propagation to derive a confounder-free saliency map. }
\label{fig-1}
\end{figure}

\subsection{Univariate Test for Identifying Confounding Effects}
This section introduces a way to test for the presence of confounding effect within a specific feature. Let $\boldsymbol{f}^j=[f_1^j,...,f_N^j]$ denote the $j^{th}$ feature derived from all $N$ training images. Likewise, denote $\boldsymbol{s}=[s_1,...,s_N]$ as the $N$ prediction scores and $\boldsymbol{z}=[z_1,...,z_N]$ as a confounding variable (e.g., age of the $N$ subjects). In this work, we use GLM \cite{dobson1990glm} to perform a group-level statistical test for detecting whether the relationship between $\boldsymbol{s}$ and $\boldsymbol{f}^j$ is confounded by $\boldsymbol{z}$. Specifically, GLM decomposes the variance in $\boldsymbol{f}^j$ into variance explained by $\boldsymbol{s}$ and variance explained by $\boldsymbol{z}$. The model reads
\begin{equation}
    \boldsymbol{f}^j = \beta_0 + \beta_1 \boldsymbol{s} + \beta_2 \boldsymbol{z}.
    \label{eq:glm}
\end{equation}
We claim feature $\boldsymbol{f}^j$ is confounded by $\boldsymbol{z}$ if the null hypothesis that linear coefficient $\beta_2$ is zero can be rejected (e.g., $p<0.05$ by \textit{t}-test). In other words, when the variance in $\boldsymbol{f}^j$ is partially explained by $\boldsymbol{z}$, $\boldsymbol{f}^j$ potentially contributes to the prediction of the confounder rather than the key variable of interest. This analysis can be extended to handle multiple confounding variables, where all confounders are included in the GLM as independent covariates. Then, $\boldsymbol{f}^j$ is confounded when the $p$-value for at least one confounder is significant. Note, this model is a specific instance of the mediation model \cite{MacKinnon2008}, a popular model for confounding analysis. However, our model makes fewer assumptions so that it is more sensitive in detecting confounding effects than the mediation model. We also emphasize that such confounding analysis can only be performed on the feature-level instead of voxel-level. Unlike features encoding geometric patterns that are commensurate within a group, voxel intensities are only meaningful within a neighborhood but variant across MRIs. As such, removing confounding effects based on feature-analysis is prevalent in traditional feature-based models (non-deep-learning models) \cite{Adeli2018,park2018}.

Repeating the above analysis for all $M$ features, we generate a binary mask $\boldsymbol{b}\in [0,1]^M=[b^1,...,b^M]$, where $b^j=0$ indicates the presence of confounding effect in the $j^{th}$ feature and $b_j=1$ otherwise. 

\subsection{Visualization via Partial Back-Propagation}
To generate a saliency map unbiased towards the subset of confounded features, we further investigate the voxelwise partial derivative. Based on the chain-rule,
\begin{equation}
    \frac{\partial s_i}{\partial \mathcal{I}_v}=\frac{\partial s_i(f_i^1,...,f_i^M)}{\partial \mathcal{I}_v}=\sum_{j=1}^M \frac{\partial s_i}{\partial f_i^j} \frac{\partial f_i^j}{\partial \mathcal{I}_v}.
    \label{eq:chain_rule}
\end{equation}
Eq. \eqref{eq:chain_rule} factorizes the voxelwise partial derivative with respect to the $M$ features, where each $\partial s_i / \partial f_i^j$ quantifies the impact of the $j^{th}$ feature on the prediction. Therefore, to derive a confounder-free saliency map, we set this impact to zero for the confounded features. In doing so, the saliency score can be computed as  
\begin{equation}
    \sum_{j=1}^M b^j \frac{\partial s_i}{\partial f_i^j} \frac{\partial f_i^j}{\partial \mathcal{I}_v}.
    \label{eq:partial_bp}
\end{equation}
Computationally, this corresponds to a partial back-propagation procedure, where the gradient is only back-propagated through the un-confounded features. \bigbreak

\noindent\textbf{The Refactorization Trick.} We show that performing the partial back-propagation for a training image $\mathcal{I}$ can be implemented by refactorizing the trained ConvNet model and then applying the original visualization pipeline of full back-propagation. As enforcing a zero $\partial f_i^j / \partial \mathcal{I}_v$ is equivalent to fixing $f_i^j$ to a constant value independent of the input image, we design a dummy layer $\mathcal{L}$ between the encoder and the predictor that performs $\mathcal{L}(\boldsymbol{x}) = \boldsymbol{x}\otimes \boldsymbol{b}_i \oplus ((1-\boldsymbol{b}_i) \otimes \boldsymbol{y}_i)$, where $\otimes$ and $\oplus$ denote element-wise operators, and $\boldsymbol{y}_i$ is a constant feature vector for image $i$ pre-computed by the trained ConvNet. As shown in Fig. \ref{fig-1}b, the dummy layer fixes the value of confounded features while keeping un-confounded features dependent on the input image. As such, the partial back-propagation of Eq. \eqref{eq:partial_bp} can by simply computed by running the full back-propagation on the refactorized model. Note, model refactorization is performed for each MR image independently to yield subject-specific saliency maps.

\section{Experiments}
We first performed synthetic experiments, in which image data were imputed by known confounding effects so that we could test whether the proposed approach can successfully remove those effects during visualization. Next, we applied the approach to two real datasets to visualize (1) the impact of HIV on brain structures while controlling for aging effects; (2) sexual dimorphism during adolescence while controlling for `puberty stage'.

\begin{figure}[!t]
\centering
\includegraphics[width=\linewidth]{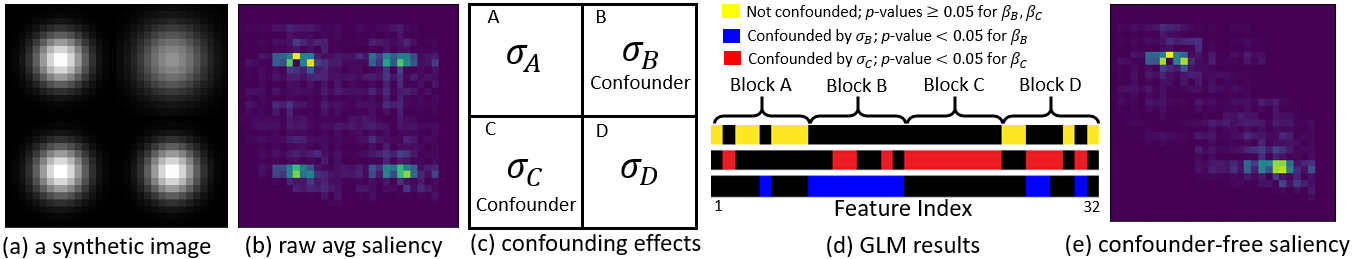}
\caption{Synthetic experiments: (a) Each synthetic image contains 4 Gaussians that are created differently between the two groups; (b) Average saliency map produced by the original visualization pipeline; (c) Widths of the two off-diagonal Gaussians are considered as confounders; (d) GLM identifies selective features, mainly in Blocks B and C, as confounded; (e) Removing the confounded features in the visualization leads to a confounder-free saliency map. }
\label{fig-2}
\end{figure}

\subsection{Synthetic Data}
We first generated a synthetic dataset containing two groups. Each group consisted of 512 2D images (dimension: $32\times32$ pixels). Each image was generated by 4 Gaussians (Fig. \ref{fig-2}a), the width of which was controlled by the standard deviation $\sigma$. For each image of Group 1, we sampled $\sigma$ from the uniform distribution $\mathcal{U}(2,6)$. Images of Group 2 generally had wider distributions as we sampled from $\mathcal{U}(4,8)$ instead. To predict group labels from the synthetic images, we constructed a simple ConvNet with the encoder consisting of 3 stacks of 2*2 convolution/ReLu/max-pooling layers and producing 32 intermediate features. The fully-connected predictor had one hidden layer of dimension 16 with \texttt{tanh} as the non-linear activation function. We trained the network for binary classification on the entire synthetic dataset as the focus here was to interpret the trained model as opposed to measuring classification accuracy. With the trained ConvNet, we first applied the original visualization pipeline to each image and averaged the resulting subject-specific saliency maps. The average saliency map shown in Fig. \ref{fig-2}b indicates that all 4 Gaussians contributed to the classification.

Next, we viewed the width of the two off-diagonal Gaussians, i.e., the standard deviations $\boldsymbol{\sigma}_B$ of Block B and $\boldsymbol{\sigma}_C$ of Block C as confounders. Based on Eq. \ref{eq:glm}, we then tested the presence of confounding effects in each of the 32 intermediate features with the following GLM:
$\boldsymbol{f}^j = \beta_0 + \beta_1 \boldsymbol{s} + \beta_B \boldsymbol{\sigma}_B + \beta_C \boldsymbol{\sigma}_C$.
The results revealed that all features extracted from Blocks B and C were detected as confounded ($p<0.05$ for either $\beta_B$ or $\beta_C$), while only features from Blocks A and D were identified as unconfounded ($p\geq0.05$ for both $\beta_B$ and $\beta_C$). We can see that our conservative test was sensitive in detecting confounding effects (no false negative but several false positives), thereby potentially removing some features representing true group difference. Such trade-off can be controlled by the $p$-value threshold used in the GLM tests. Finally, using the binary mask (yellow mask in Fig. \ref{fig-2}d) for partial back-propagation, we produced a confounder-free average saliency map (Fig. \ref{fig-2}e) that successfully removed the confounding effects.

\subsection{Visualizing HIV Effects}
The second experiment examined the impact of HIV on the human brain. The classification was performed on the T1-weighted MRI data of 223 control subjects (CTRL) and 122 HIV patients \cite{Adeli2018}. Participants ranged in age between 18–86 years, and there was a significant age difference between CTRL and HIV subjects (CTRL: $45\pm17$, HIV: $51\pm8.3$, \textit{p}<0.001 by two-sample $t$-test). As HIV has been frequently suggested to accelerate brain aging \cite{cole2017increased}, age is therefore a confounder that needs to be controlled for when interpreting the saliency map associated with the trained classifier. \bigskip

\noindent\textbf{Preprocessing and Classification.} The MR images were first preprocessed by denoising, bias field correction, skull striping, affine registration to the SRI24 template (which accounts for differences in head size), and re-scaling to a $64\times64\times64$ volume \cite{Adeli2018}. Even though the present study focused on the visualization technique, we measured the classification accuracy as a sanity check via 5-fold cross validation. To ensure the classifier can reasonably learn the group difference between HIV and CTRL subjects, the training dataset was augmented by random shifting (within one-voxel distance), rotation (within one degree) in all 3 directions, and left-right flipping. Note, the flipping was based on the assumption that  HIV infection affects the brain bilaterally \cite{Adeli2018}. The data augmentation resulted in a balanced training set of 1024 CTRLs and 1024 HIVs.

As the flipping removed left-right orientation, the ConvNet was built on half of the 3D volume containing one hemisphere. The encoder contained 4 stacks of 2*2*2 3D convolution/ReLu/batch-normalization/max-pooling layers yielding 4096 intermediate features. The fully-connected predictor had 2 hidden layers of dimension (64, 32) with \texttt{tanh} as the non-linear activation function. An L2-regularization ($\lambda = 0.1$) was applied to all fully-connected layers. Based on this ConvNet architecture, we achieved 73\% normalized accuracy for HIV/CTRL classification, which was comparable to other recent studies on this dataset \cite{Adeli2018}. \bigskip

\begin{figure}[!t]
\centering
\includegraphics[width=\linewidth]{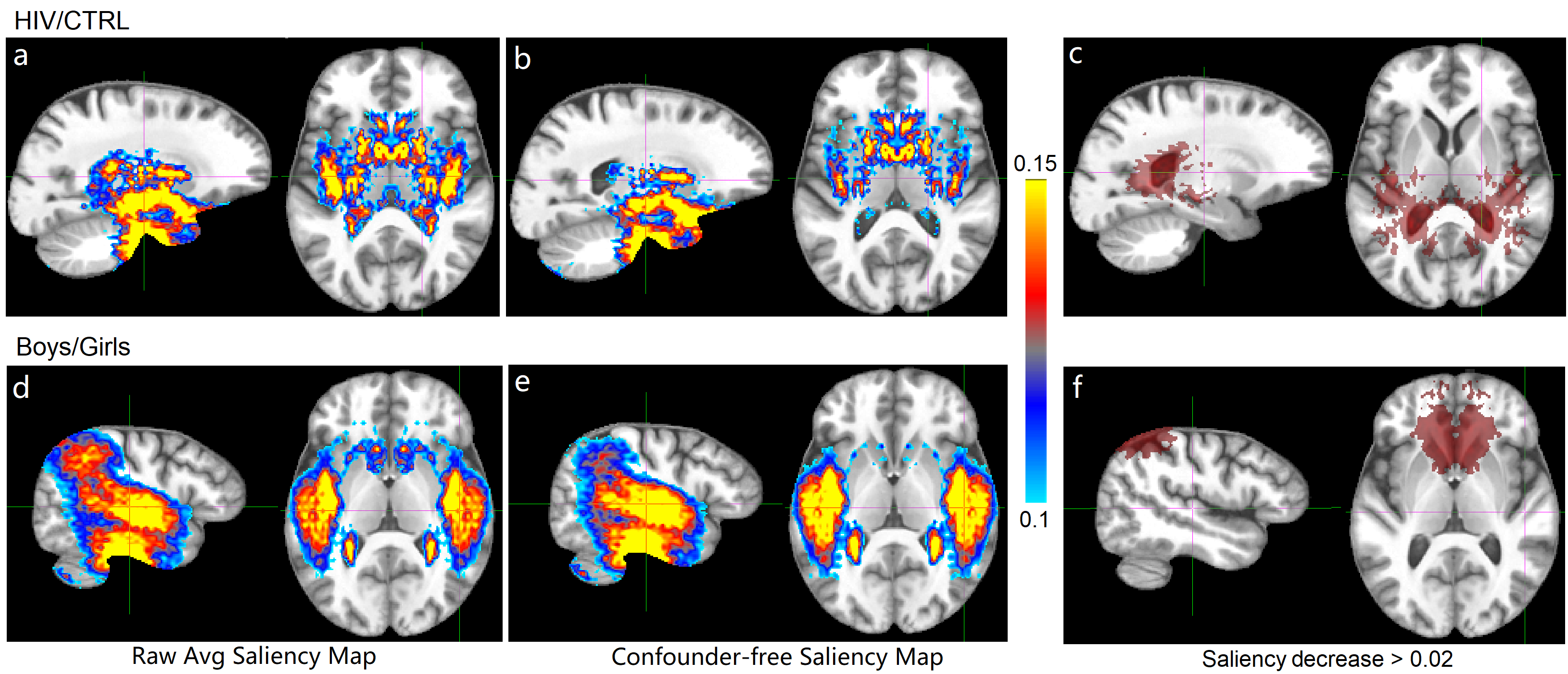}
\caption{Visualization of ConvNets trained for HIV/CTRL classification (top row) and sexual dimorphism (bottom row). }
\label{fig-3}
\end{figure}

\noindent\textbf{Model Visualization.} To visualize the HIV effect, we re-trained the ConvNet on a dataset of 1024 CTRLs and 1024 HIVs augmented from the entire dataset of 345 MRIs. We first visualized the average saliency map produced by the original visualization pipeline. Since the ConvNet operated on only one hemisphere, we mirrored the resulting average saliency map to the other hemisphere to create bilaterally symmetric display and overlaid it on the SRI24 T1 atlas (Fig. \ref{fig-3}a). For comparison, we then visualized the confounder-free saliency map produced by our approach. Specifically, we tested each of the 4096 features with $\boldsymbol{f}^j = \beta_0 + \beta_1 \boldsymbol{s} + \beta_2 age$, and 804 were identified to be confounded by age. Fig. \ref{fig-3}b shows the saliency map after removing aging effects, and Fig. \ref{fig-3}c shows that saliency at the posterior ventricle (red regions) was attenuated by our approach indicating those regions contained aging effects instead of HIV effects. This finding is consistent with current concept that the ventricular volume significantly increases with age \cite{Kaye92}.

\subsection{Visualizing Sexual Dimorphism}
The third experiment aimed to improve understanding of sexual dimorphism in brain development that emerges during adolescence. The classification was performed on the baseline T1 MR images of 334 boys and 340 girls (age 12-21) from the National Consortium on Alocohol and NeuroDevelopment in Adolescence (NCANDA) \cite{Brown2015}. All subjects met the no-to-low alcohol drinking criteria of the study, and there was no significant age-difference between boys and girls ($p$>0.5 two-sample $t$-test). As puberty stage \cite{Brown2015} of girls was significantly higher than boys during adolescence, the pubertal development score (PDS: boys 2.86$\pm$0.7, girls 3.41$\pm$0.6, \textit{p}<0.001 by two-sample $t$-test) was a potential confounder of the study.

All experimental setups complied with the previous HIV study. As the first attempt of predicting sex on the NCANDA data, we achieved 89.5\% normalized accuracy based on a 5-fold cross-validation. The original saliency map produced for the ConvNet trained on the entire augmented dataset is shown in Fig. \ref{fig-3}d. After testing and removing PDS effects, the confounder-free saliency map is shown in Fig. \ref{fig-3}e. Consistent with existing adolescence literature, sex difference was mainly found in the temporal lobe \cite{Sowell2002}. Fig. \ref{fig-3}f indicates PDS effects mainly existed in the frontal and inferior parietal region. Another interesting observation is in the caudate, which has been frequently reported as proportionately larger in female participants across different ages \cite{MacKinnon2008}. As shown in our results, the saliency at the caudate region attenuated after removing confounding effects, suggesting a potential compounding effect of PDS in that region.

\section{Discussion and Conclusion}
In this paper, we introduced a novel approach for confounder-free visualization and interpretation of a trained ConvNet. By performing partial back-propagation with respect to a set of unconfounded intermediate features, the approach disentangled true group difference from confounding effects and produced unbiased saliency maps. We successfully illustrated its usage on a synthetic dataset with ground-truth confounding effects and two real neuroimaging datasets. Because our approach is a type of post-hoc analyses with respect to a trained model, further extension could potentially integrate similar confounder-control
procedures during model-training time to fully explore unbiased group differences
within a dataset.

\noindent\textbf{Acknowledgements.} This research was supported in part by NIH grants AA017347, AA005965, AA010723, AA021697, AA013521, AA026762 and MH113406.

\bibliographystyle{splncs}
{ \small
\bibliography{refs}
}

%
%
%
%
%
%
\end{document}